\documentclass[letterpaper]{article} 
\usepackage{aaai24}  
\usepackage{times}  
\usepackage{helvet}  
\usepackage{courier}  
\usepackage[hyphens]{url}  
\usepackage{graphicx} 
\usepackage{natbib}  
\usepackage{caption} 
\usepackage{algorithm}
\usepackage{algorithmic}

\usepackage{newfloat}
\usepackage{listings}
\DeclareCaptionStyle{ruled}{labelfont=normalfont,labelsep=colon,strut=off} 
\floatstyle{ruled}
\newfloat{listing}{tb}{lst}{}
\floatname{listing}{Listing}
\usepackage{color}
\usepackage{booktabs}
\usepackage{threeparttable}
\usepackage{multirow}
\usepackage{amsmath}

\title{Towards Equipping Transformer with the Ability of Systematic Compositionality}
\author{
    Chen Huang, Peixin Qin, Wenqiang Lei\thanks{Correspondence to Wenqiang Lei.}, Jiancheng Lv
}
\affiliations{
    College of Computer Science, Sichuan University, Chengdu, China\\
    Engineering Research Center of Machine Learning and Industry Intelligence, Ministry of Education, Chengdu, China\\
    \{huangc.scu, qinpeixin.scu, wenqianglei\}@gmail.com
}

\usepackage{bibentry}

\begin{document}

\maketitle

\begin{abstract}
One of the key factors in language productivity and human cognition is the ability of \textit{systematic compositionality}, which refers to understanding composed unseen examples of seen primitives. However, recent evidence reveals that the Transformers have difficulty generalizing the composed context based on the seen primitives.
To this end, we take the first step to propose a compositionality-aware Transformer called CAT and two novel pre-training tasks to facilitate systematic compositionality. We tentatively provide a successful implementation of a multi-layer CAT on the basis of the especially popular BERT.
The experimental results demonstrate that CAT outperforms baselines on compositionality-aware tasks with minimal impact on the effectiveness on standardized language understanding tasks.
\end{abstract}

\section{Introduction}
Linguistic research confirms the discreteness of linguistic symbols and their compositionality to construct larger linguistic expressions \cite{montague1970universal, frege1948sense, baroni2020linguistic, akyurek2022compositionality}. These characteristics are known as \textit{Systematic Compositionality} \cite{fodor1988connectionism, keysers2019measuring, lake2017building}. For instance, sentences are built from words and phrases.  
Such systematic compositionality fosters humans ability to understand and generalize to unseen combinations of seen primitives \cite{lake2017building} and model complex phenomena \cite{liu2021discrete}.
Therefore, it is widely recognized as an essential capability of human intelligence \cite{ma2023brain}.

However, previous studies have shown that language models struggle with generalizing through composition \cite{cartuyvels2021discrete, lake2018generalization, loula2018rearranging}. Even for large language models (LLMs), recent evidence suggests that they still struggle to establish systematic compositionality after fine-tuning on compositionality-aware datasets \cite{yu2021interplay} or prompting with in-context examples \cite{an2023context}. 
The challenge lies in the fact that the semantics of a group of primitives vary depending on their meanings and how they are combined.
Considering examples on word-level compositionality in Table \ref{tab:example} where words serve as the primitives of phrases, two phrases composed of different words may have similar semantics (e.g., \textit{Safety officer} and \textit{Security guard}), while phrases may exhibit different semantics through different combinations (e.g., \textit{Water body} and \textit{Body water}). Empirical evidence \footnote{See Table \ref{tab:example} and Appendix \ref{asd8fonq34wrf} for details.} shows that even the 'omnipotent' ChatGPT still struggles to accurately capture the semantic changes among different word combinations, failing to align with human judgments regarding the similarity of phrase pairs. 
One possible explanation is that despite the ability to capture the meaning of words, current Transformer frameworks fail to develop systematic compositional skills \cite{dziri2023faith, ma2023brain}.
In contrast, humans certainly do understand language by learning the meaning of words and composing more elaborate meanings \cite{cartuyvels2021discrete}. Therefore, to achieve human-level language understanding, it is imperative to invest more effort in building a compositionality-aware model that promotes LLMs with stronger capabilities in systematic compositionality.

\begin{table}[t]
\centering
\caption{Compositionality example from \citet{asaadi-etal-2019-big}. ChatGPT struggles to achieve human-level language understanding as it fails to align with human judgments regarding the similarity of phrase pairs that are formed by different word composition. 
}
\label{tab:example}
\resizebox{0.38\textwidth}{!}{%
\begin{tabular}{c|c|c|c}
\toprule
\multicolumn{1}{c|}{} & \multicolumn{1}{c|}{} & \multicolumn{2}{c}{\textbf{Similarity Score}} \\ \cline{3-4} 
\multicolumn{1}{c|}{\multirow{-2}{*}{\textbf{Phrase 1}}} & \multicolumn{1}{c|}{\multirow{-2}{*}{\textbf{Phrase 2}}} & \multicolumn{1}{c}{\textbf{LLM}} & \multicolumn{1}{c}{\textbf{Human}} \\ \midrule
Safety officer & Security guard & 0.6 & 0.881 \\ \midrule
Water body & Body water & 0.7 & 0.381 \\ \bottomrule
\end{tabular}%
}
\end{table}

\begin{figure}[!htb]
   \centering  
   \includegraphics[width=0.45\textwidth]{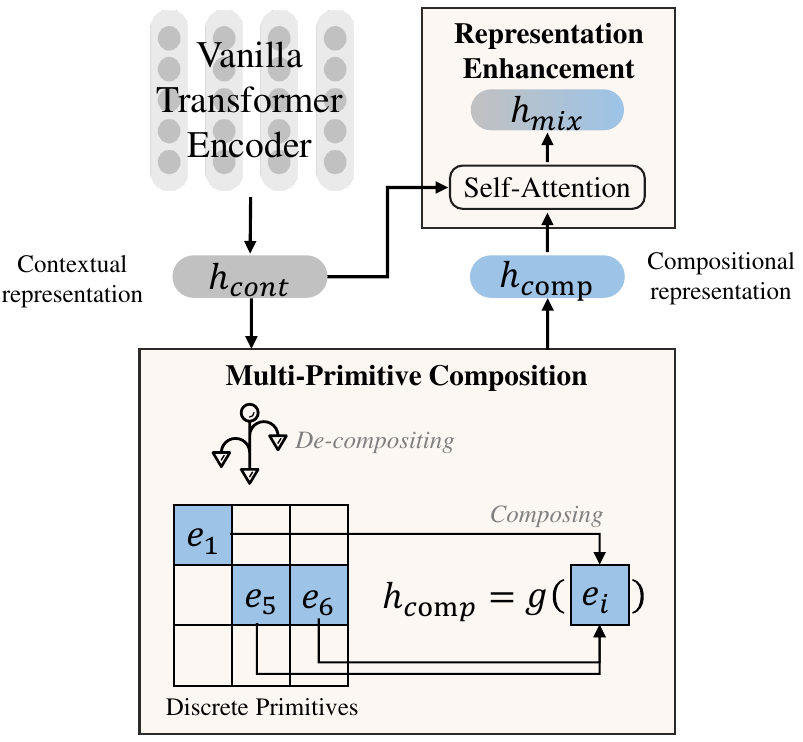} 
   \caption{Illustration of CAT, which contains two modules. By learning to compose primitives, CAT facilitates systematic compositionality of the vanilla Transformer.}  
   \label{fig:demograph}
\end{figure}

Motivated by this, we take the first step to propose a \underline{C}ompositionality-\underline{A}ware \underline{T}ransformer called \textbf{CAT} to facilitate the systematic compositionality, along with two novel pre-training tasks. 
1) As depicted in Fig.~\ref{fig:demograph}, CAT introduces two modules: Multi-Primitive Composition and Representation Enhancement, to the vanilla Transformer encoder. These modules enable CAT to learn how to compose primitives and enhance the representation of the vanilla Transformer encoder, respectively. Specifically,
the Multi-Primitive Composition module decomposes a contextual word representation\footnote{Here, a word representation is the average of its corresponding token embeddings.} $h_{cont}$ into multiple primitives, which are represented as discrete latent space vectors. It then produces a compositional representation $h_{comp}$.
The Representation Enhancement module integrates $h_{comp}$ and $h_{cont}$ to yield the final output $h_{mix}$ with stronger systematic compositionality and without losing contextual information of $h_{cont}$ required in downstream tasks.
Notably, these modules can also be applied to the [CLS] token to achieve the semantic composition of a sentence.
2) Additionally, we propose two new pre-training tasks, i.e., Guided Decomposition and Semantics Composition, to further enhance the systematic compositionality. 
The former supervises the decomposition of $h_{cont}$ using the OpenHowNet dataset \cite{qi2019openhownet}, which records the mappings from words to their corresponding primitives (i.e., Sememes\footnote{Minimum semantic units of our languages. A  set of discrete sememes (such as \textit{Family}, \textit{Spouse}, \textit{Female}) could compose the meanings of all the words (\textit{Wife}, in this case) \cite{qi2019modeling}.}). 
The latter guides the composition of discrete primitives so that $h_{comp}$ and $h_{mix}$ are semantically informative without losing contextual information required in the downstream tasks.
As such, by learning to compose primitives during pre-training, CAT could facilitate systematic compositionality of the vanilla Transformer.

As suggested in previous studies \cite{yu2021interplay, cartuyvels2021discrete, hendrycks2020pretrained, yu2020assessing}, our evaluation closely revolves around the characteristics of the systematic compositionality to provide detailed insights. Due to the issue of computational resources, we restrict our analysis and comparison to the widely used BERT and tentatively implement and pre-train a multi-layer CAT from scratch following BERT. 
Our experimental findings demonstrate that our approach outperforms BERT and other baselines on compositionality-aware tasks while having minimal impact on the effectiveness of standardized language understanding tasks. In comparison to BERT, CAT exhibits superior performance in identifying semantic changes in both phrase-level and sentence-level compositionality-aware tasks without losing any advantages in standardized GLUE tasks. On average, our method experiences performance gains of +6.42 and +1.10, respectively. Furthermore, CAT shows promising improvements in compositional generalization (+1.83) and robustness to noisy context (+3.09).
Given that compositionality has always been considered a major factor in language productivity and human cognition, we believe our work is an important proof-of-concept for promoting LLMs' capability in systematic compositionality. In conclusion, we claim the following contributions.
\begin{itemize}
    \item For the first time, we propose a systematic compositionality aware Transformer called CAT, which explicitly equips the vanilla Transformer with the ability to learn to compose primitives.
    \item We also propose two novel pre-training tasks to further facilitate the systematic compositionality of CAT.
    \item We verify our effectiveness with extensive empirical studies and offer an in-depth analysis. We provide insight for future studies on LLM's systematic compositionality. 
\end{itemize}

\section{Related Work}
Our focus is on the discreteness of linguistic symbols and their systematic compositionality. Therefore, we conduct a literature review on systematic compositionality and existing techniques for discretization \& compositionality.

\textbf{Systematic Compositionality}. Systematic compositionality in the language is an attractive capability of combining discrete elementary units in systematic ways to create compound ones \cite{montague1970universal, frege1948sense, baroni2020linguistic, akyurek2022compositionality}.
It allows humans to make so-called “infinite use of finite means” \cite{chomsky2014aspects} and fosters the capacity of generalization \cite{lake2017building}. 
It has led to improved performance in various NLP tasks such as question answering \cite{bogin2021latent} and machine translation \cite{ataman2018compositional}. In the era of LLMs, however, recent evidence shows that fine-tuning a given pre-trained model on a specific task may not improve its compositionality capability \cite{yu2021interplay}. Over-parameterized LLMs, like ChatGPT, are still sensitive to the selection of in-context examples \cite{an2023context}, making their compositionality ability fragile. More recently, researchers have found that current Transformers fail to develop systematic problem-solving skills \cite{dziri2023faith, ma2023brain}, which highlights the need for a compositionality-aware Transformer that may enhance LLMs' capabilities in systematic compositionality. To this end, we take the first step to propose a compositionality-aware Transformer and two novel pre-training tasks to facilitate systematic compositionality.

\textbf{Techniques for Discretization \& Compositionality}. Exploring the compositional characteristics has drawn lots of attention, but they focus on composing continuous primitives. Notably, a continuous primitive is a continuous, real-valued variable that takes on values in connected regions of $R^n$. while a discrete primitive is a discretely valued variable that takes on either a limited or a countably infinite number of distinct values \cite{cartuyvels2021discrete}. To compose continuous primitives, existing techniques include the group-equivariance theory \cite{higgins2018towards, gordon2019permutation}, syntactic attention \cite{russin2019compositional, li2019compositional}, disentangled representation \cite{burgess2018understanding, locatello2019challenging}, data argumentation \cite{akyurek2022compositionality, andreas2020good}, or neural modularization \cite{andreas2016neural}. 
However, considering the discreteness of linguistic symbols, research on compositionality with discrete primitives is relatively rare.
One possible reason for this is that optimizing discrete primitives during the backpropagation of a neural network in a stabilized and bias-free way is more challenging than optimizing contextual ones \cite{friede2021efficient, farinhas2021sparse}. Recent vector quantization techniques \cite{van2017neural, liu2021discrete} propose using the Straight-Through-Estimator \cite{courbariaux2015binaryconnect} to facilitate optimization on discrete codes/primitives. Nevertheless, the composition of discrete primitives is yet to be explored, let alone the pre-trained language model with systematic compositionality. 
As a result, it is unclear how to create a compositionality-aware language model and assess its performance on more general tasks.

\section{Compositionality-Aware Transformer}
 


To promote systematic compositionality, CAT introduces two modules, i.e., Multi-Primitive Composition and Representation Enhancement, to the vanilla Transformer encoder. These modules enable CAT to learn how to compose primitives and enhance the representation of the vanilla Transformer encoder, respectively.



\subsection{Multi-Primitive Composition Module}
This module explicitly equips the vanilla Transformer encoder with an ability of learning to compose primitives. It decomposes a contextual word representation $h_{cont}$ into multiple primitives that bear the strong similarities. It then produces a compositional representation $h_{comp}$.

\textbf{Primitives Representation}. For each $h_{cont}$, we assume that they are grounded in the same semantic space\footnote{This is inspired by \textit{mutual knowledge hypothesis} \cite{sperber1986relevance}, saying that knowledge required to interpret a message is grounded in the understanding of the message sender and receiver.}, spanned by a limited number of distinct primitives. This forces the model to decode the elementary units from the contextual word representation $h_{cont}$. Considering the discreteness of language symbols, we require the $i$-th primitive to be represented by a discrete latent space vector $e_i \in R^m$, where $m$ is the dimension size. Denoting $K$ as the size of the discrete latent space, all the semantic space in CAT is grounded in an $L$-way categorical variable, $e \in R^{K\times m}$, which we refer to as a \textit{codebook}. For instance,  Fig.\ref{fig:demograph} contains a codebook with nine discrete primitives/codes. Note that the codebook is a trainable parameter.

\textbf{Decompositing into Primitives}. 
Given a contextual representation $h_{cont}$, one way to decompose it is the vector quantization \cite{van2017neural, razavi2019generating, liu2021discrete, liu2022adaptive}, which involves learning a discrete latent representation for an input vector. Given an input vector $h_{cont} \in R^m$, the vector quantization method maps $h_{cont}$ to the nearest-neighbor quantized code in the codebook $e^{K \times m}$. More concretely, the discretization process for vector $h_{cont}$ is described as follows.
\begin{equation}
\begin{split}
    e_{o_j}&=\text{Discretize}(h_{cont}),
    \\\text{where } o_i&=\mathop{\arg\min}_{j \in \{1,...,K\}}\| h_{cont} - e_j \|^2_2,
\end{split}
\end{equation}
where $e_j$ is $j$-th code in the codebook $e$. Finally, the contextual representation $h_{cont}$ is quantified using 'hard K-Means clustering' and discretized into one code. 

However, his approach fails to meet the requirements for decomposition as it only produces one code. Also, the fitness of one code is limited compared to the original semantics of $h_{cont}$. To address this, our \textit{multi-primitive composition module} decomposes a contextual representation into multiple discrete codes. In particular, our method uses a soft version of K-means clustering, where the number of clusters is dynamically learned. We achieve this by replacing the $\mathop{\arg\min}$ operator with the sparse attention mechanism \cite{zhang2021sparse} as follows.
\begin{equation}
    O=\text{RMSNorm}(\text{ReLU}(f(Q, K))),
\end{equation}
where $Q=h_{cont}W_Q$, $K=eW_K$, and $f$ is a scoring function like cosine. The vector $O$ represents the sparse attention weight, where each index corresponds to a code in the codebook $e$. The RMSNorm operator also is utilized to increase the optimization stabilization, as suggested in \citet{zhang2021sparse}.
By this means, our multi-primitive composition filters out irrelevant codes from the compositions of $h_{comp}$ and allows us to decompose $h_{cont}$ into multiple latent codes of dynamic size.
It is worth mentioning that our multi-primitive composition has the additional benefit of addressing the differentiation challenge. Unlike raw vector quantization, which approximates the gradient of the $\mathop{\arg\min}$ operator using Straight-Through-Estimator \cite{courbariaux2015binaryconnect} or Gumbel-Softmax \cite{jang2016categorical}, both of which can be unstable \cite{yin2018understanding}, our approach is entirely differentiable.

\textbf{Composing Primitives}. To capture the semantics of various combinations, it is important that different primitives have varying significance in the composition process, depending on the context. We accomplish this by utilizing the attention vector $O \in R^K$ to select multiple discrete primitives and combining them into a compositional representation $h_{comp}$. This is achieved through the function $g(O,e) = OV$, where $V=eW_V$. It is noteworthy that CAT takes into account the systematic compositionality of a sentence by breaking down the $h_{cont}$ of the [CLS] token, which is known to capture the semantics of the entire sentence.

\subsection{Representation Enhancement Module}
The objective of the Representation Enhancement module is to integrate $h_{comp}$ and $h_{cont}$ in a way that produces the ultimate output $h_{mix}$, which possesses superior systematic compositionality while retaining the contextual information of $h_{cont}$ necessary for downstream tasks. As a result, CAT enhances the representation of the vanilla Transformer $h_{cont}$.
To accomplish this, a self-attention mechanism is employed to automatically adjust the importance weights of $h_{cont}$ and $h_{comp}$ during the integration.


Formally, given an input sentence $x$ of length $N$, $i$-th word $x_i$ with its representations $h_{cont}^i$ and $h_{comp}^i$, we derive two sequence embeddings $H_{comp}=\{h_{comp}^0,h_{comp}^1, ..., h_{comp}^N\}$ and $H_{cont}=\{h_{cont}^0,h_{cont}^1, ..., h_{cont}^N\}$, $h_{cont}^0$ and $h_{comp}^0$ correspond to representations of [CLS]. To integrate them, we feed the concatenatation of $H_{comp}$ and $H_{cont}$ into a self-attention layer to obtain the final outputs $H=\{h_{mix}^0, h_{mix}^1, ..., h_{mix}^N\}$, where $h_mix^i$ is the mixed representation of $x_i$ or [CLS]. This process is described as follows, where the second step re-scales the dimension for multi-layer CAT when necessary. 
\begin{equation}
    \begin{split}
        H &= \text{self-attention}([h_{comp}, h_{cont}])\\
        H &= \frac{1}{2}(H[:, :N, :] + H[:, N:, :]) 
    \end{split}
\end{equation}

By this means, the CAT learns to automatically fuse the contextual and compositional representations of each word according to the input context. As a result, three types of representations are generated, namely the raw contextual representations $h_{cont}$, discrete representations $h_{comp}$, and mixed representations $H$. The learned mixed representation $H$ can be used as the contextual input for the next CAT layer.



\section{Compositionality-aware Pre-training Tasks}
This section introduces two new pre-training tasks that aim to improve the systematic compositionality ability. The first task is the guided decomposition task, which utilizes sememes to supervise the semantic decomposition of $h_{cont}$. The second task is the semantics composition task, which ensures the composition of discrete primitives so that $h_{comp}$ and $h_{mix}$ are semantically informative without losing contextual information required in the downstream tasks.

\subsection{Guided Decomposition Task}
To further enhance the interpretability and learning efficiency of the multi-primitive composition module, we resort to the semantic compositionality \cite{qi2019modeling, wierzbicka1996semantics} and utilize sememes, which are defined as the minimum semantic units of human languages \cite{bloomfield1926set}, to supervise the decomposition. Building upon the assumption that the meanings of all the words can be composed of a limited set of sememes \cite{qi2019modeling, wierzbicka1996semantics}, each code/primitive in the CAT's codebook is defined as a specific sememe. Thus, the compositional representation can be interpreted as the semantic composition of selected sememes. Consequently, the goal of this task is to learn the map from each word to its corresponding sememes.

\textbf{Sememes Guided Decomposition}. To achieve this, we resort to the OpenHowNet \cite{qi2019openhownet}, a widely acknowledged sememe knowledge base, as the supervision of semantic decomposition. We require the sparse attention mechanism to attend to the appropriate sememes in the codebook. Here, $x_i$ represents the $i$-th word in an input sentence of length $N$, $e \in R^{K \times m}$ is the codebook in CAT (or the last layer of the multi-layer CAT), $O_i \in R^K$ is the attention weight to the codebook and $s_i \in R^K$ is a $\{0,1\}$ vector containing the correct sememes for $x_i$, where $s_{ij}=1$ if the $j$-th sememe belongs to $x_i$, otherwise $s_{ij}=0$. To calculate the loss $L_{GD}$ for incorrect sememe-matchings, we averaged the attention weights that are focused on the wrong sememes. Furthermore, we introduced the L1 norm on the attention weights from all layers to encourage sparsity.
\begin{equation}
    L_{GD}(x_i) = \sum_{x_i \in x}\frac{sum((1-s_i)\odot O_i)}{sum(1-s_i)} + L_1(O_i)
\end{equation}

\subsection{Semantics Composition Task}
The goal of the semantics composition task is to encourage $h_{comp}$ and $h_{mix}$ to be semantically informative without losing contextual information required in the downstream tasks. Here, we involve the following three learning objectives.

\textbf{Reconstruction Consistency}. It aims to encourage the compositional representations $h_{comp}$ to bear strong similarity with the contextual ones $h_{cont}$ in the vector space. We consider the following objectives, where the stop gradient operator $sg$ and hyper-parameter $\beta \in [0,1)$ are utilized to shift more optimization focus on updating discrete representation, rather than the contextual one. 
    \begin{equation}
        \ell_{rc} = \| H_{comp} - sg(H_{cont}) \|_2^2 + \beta \| sg(H_{comp}) - H_{cont} \|_2^2
    \end{equation}

\textbf{Semantic Sufficiency}. It concerns that $h_{cont}$, $h_{comp}$, and $h_{mix}$ contain sufficient semantic information. They should be effective on the pre-training task $\text{Ptask}$ used in BERT.
    \begin{equation}
        \ell_{ss} = \text{Ptask}(H_{comp}) + \text{Ptask}(H_{cont}) + \text{Ptask}(H)
    \end{equation}

\textbf{Nuance Minimization}. It encourages the nuance between contextual and compositional representations $H_{nu} = H_{cont} - H_{comp}$ should be less informative, insignificant to the main semantics. Thus, we aim at maximizing the entropy of $H_{nu}$ and minimizing its performance on the pre-training tasks used in the semantic sufficiency.
    \begin{equation}
      \ell_{nm} =  exp(-\text{Entropy}(H_{nu})) + exp(-\text{Ptask}(H_{nu}))
    \end{equation}

Finally, we summarize the overall loss function for the semantics composition task as $L_{SC}=\ell_{rc} + \ell_{ss} + \ell_{nm}$. The overall loss function for CAT pre-training is $L = L_{task} + L_{GD} + L_{SC}$, where $L_{task}$ is the loss for other pre-training tasks used in BERT.

\section{Experiments}
In this section, we concentrate on comparing our approach to the vanilla Transformer and conducting comprehensive evaluations that center on the features of systematic compositionality (as discussed in Section \ref{q1}). Additionally, we explore other characteristics of CAT (as discussed in Section \ref{q2} and \ref{q3}), including robustness and combinatorial effectiveness.

\subsection{Experimental Setup}
\textbf{Model Pre-training}. We restrict our discussions to the vanilla Transformer framework, with a special focus on the popular BERT\footnote{We restrict our discussions to BERT without loss of generality. Due to computational resource limitations, we plan to implement a model with more significant parameter quantities in the future. } to provide more detailed insights. To ensure the fairness of the experiment, we strictly follow the pre-training process of BERT \cite{kenton2019bert} in our experiments. Specifically, we implement a multi-layer CAT (MCAT) and pre-train it from scratch. The BooksCorpus (800M words), English Wikipedia (2,500M words), and extra OpenHowNet dataset serve as our pre-training data for MCAT. We utilize the pre-training tasks of BERT, namely Masked LM and Next Sentence Prediction, as the $\text{Ptask}$ in our Semantics Composition. See Appendix \ref{pre-trainingt} for more information on pre-training. 

\textbf{Baselines \& Implementation.} To simplify, we use the notations CAT$_{\text{cont}}$, CAT$_{\text{comp}}$, and CAT$_{\text{mix}}$ to represent the contextual, compositional, and mixed representations of our multi-layer CAT, respectively. As for the baselines, BERT is of particular importance to us due to its relevance to our network architecture and pre-training data. In addition to BERT, we also consider two other variants, namely RoBERTa\footnote{Different from BERT and us, RoBERTa use more pre-training datasets and customized framework.} \cite{liu2019roberta} and DistilBERT \cite{sanh2019distilbert}, following the approach of \citet{yu2021interplay, yu2020assessing}. For all experiments, we fine-tune each model with the corresponding backbone frozen and an additional MLP layer on top of the backbone for learning, which takes the $[CLS]$ embeddings as inputs. We selected the best model based on its performance on the validation set for downstream task testing. More implementation details can be found in Appendix \ref{ideital}.

\textbf{Evaluation Tasks}. It's important to note that most existing evaluation methods and datasets for systematic compositionality are only suitable for generative-based language models like SCAN \cite{higgins2018scan} and in-context prompting \cite{an2023context}. Instead, our evaluations follow recent works \cite{yu2021interplay, hendrycks2020pretrained, yu2020assessing}, which are designed for discriminative-based language models like BERT and ours. Since each evaluation task differs, we provide the details in the corresponding subsections. For more information, please refer to Appendix \ref{eeeeeeee}.

\subsection{Systematic Compositionality Evaluation}
\label{q1}
After pre-training on corpora with diverse words and their combinations, we assess the systematic compositionality of CAT on phrases and sentences. We also evaluate the compositional generalization to the out-of-distribution dataset. 

\begin{table}[]
\centering
\caption{Systematic compositionality evaluation (\%) on phrases and sentences. We \textbf{bold} the best results and \underline{underline} the second best. 'SO' means 'swap-only'}
\label{tab:system composition}
\resizebox{0.43\textwidth}{!}{%
\begin{tabular}{l|c|cc}
\toprule
\multirow{2}{*}{\textbf{Methods}} & \textbf{Adversary} & \multicolumn{2}{c}{\textbf{Correlation}} \\ \cline{2-4} 
 & PAWS (SO) & \multicolumn{1}{c|}{BiRD} & BiRD-ABBA \\ \midrule
BERT & 88.22 & \multicolumn{1}{c|}{19.50} & 1.64 \\
RoBERTa & 89.06 & \multicolumn{1}{c|}{20.32} & 3.21\\
DistilBERT & 87.44 & \multicolumn{1}{c|}{17.49} & 1.03\\ \hline
CAT$_{cont.}$ & 89.05 & \multicolumn{1}{c|}{20.19} & 1.25 \\
CAT$_{comp.}$ & \textbf{90.33} & \multicolumn{1}{c|}{\textbf{34.44}} & \textbf{7.19} \\
CAT$_{mix}$ & \underline{89.32} & \multicolumn{1}{c|}{\underline{27.72}} & \underline{6.25}  \\ \bottomrule
\end{tabular}%
}
\end{table}

\begin{table*}[t]
\centering
\caption{Evaluation on out-of-distribution datasets (\%). We report the Pearson’s correlation coefficient for the STSB, and accuracy for the other datasets.}
\label{tab:system ood}
\resizebox{0.75\textwidth}{!}{%
\begin{tabular}{l|cc|cc|cc|cc}
\toprule
\multirow{3}{*}{\textbf{Method}} & 
\multicolumn{2}{c}{\textbf{Movie}} & 
\multicolumn{2}{c}{\textbf{STSB}} & \multicolumn{2}{c}{\textbf{MNLI}} & \multicolumn{2}{c}{\textbf{AMAZON}} \\ \cline{2-9}
 & \textbf{IMDB} & \textbf{SST2} & \textbf{Images} & \textbf{MSRvid} & \textbf{Telephone} & \textbf{Letters} & \textbf{Music} & \textbf{Video} \\ \cline{2-9}
 & \textbf{IID} & \textbf{OOD} & \textbf{IID} & \textbf{OOD} & \textbf{IID} & \textbf{OOD} & \textbf{IID} & \textbf{OOD} \\ \midrule
BERT & 79.82 & 76.61 & \underline{80.94} & 88.00 & \underline{74.87} & 72.79 & 73.32 & 65.16 \\
RoBERTa & \textbf{82.09} & 75.47 & \textbf{81.31} & 88.04 & \textbf{76.29} & 72.94 & \textbf{74.30} & 66.06 \\
DistilBERT & 78.64 & 75.07 & 80.43 & 88.38 & 73.14 & 70.03 & 73.04 & 64.58 \\ \hline
CAT$_{cont.}$ & 79.34 & 76.15 & 80.05 & 88.81 & 74.31 & 72.85 & \underline{73.98} & 65.22 \\
CAT$_{comp.}$ & 77.81 & \underline{78.56} & 78.31 & \textbf{89.67} & 73.36 & \underline{73.08} & 72.41 & \underline{67.32} \\
CAT$_{mix}$ & \underline{80.23} & \textbf{79.36} & 79.97 & \underline{89.13} & \underline{74.87} & \textbf{73.96} & 73.88 & \textbf{67.44} \\ \bottomrule
\end{tabular}%
}
\end{table*}

\noindent\textbf{5.2.1 Evaluation on Phrases \& Sentences}

\textbf{Task \& Dataset}. Our goal is to assess the capability of recognizing the semantic correlation that results from different word compositionality.
In this section, we consider the two evaluation sub-tasks, i.e., phrase correlation and adversarial paraphrase sentence classification, to assess the semantics correlation of different phrases and different sentences, respectively. 
In our experiments, inspired by previous studies \cite{yu2021interplay, yu2020assessing}, we fine-tune each model on datasets that are good candidates for requiring composition and test the fine-tuned model on these two sub-tasks. For more details on the datasets and model fine-tuning, please refer to Appendix \ref{system_apa}.

\begin{itemize}
    \item \underline{Phrase Correlation}. It aims to evaluate whether CAT and baselines capture compositional phrase information and identify the semantic correlation between two phrases. Following \citet{yu2021interplay, yu2020assessing}, we fine-tune each model on the PAWS \cite{zhang2019paws}, which consists of sentence pairs with high lexical overlap. This fine-tuning task is formulated as a binary classification of whether two sentences are paraphrases or not. We then assess the fine-tuned model on BiRD \cite{asaadi-etal-2019-big}, which is a bigram-relatedness dataset designed to evaluate composition (e.g., \textit{Safety officer} and \textit{Security guard}). In addition to testing on the full BiRD dataset, we conduct a controlled experiment to remove the effects of word overlap by filtering the BiRD dataset to pairs in which the two phrases consist of the same words (e.g., \textit{Water body} and \textit{Body water}). We refer to the filtered dataset as BiRD-ABBA. For both BiRD and BiRD-ABBA, the model performance is measured by the alignment with human judgments of phrase meaning correlation. We report the Pearson correlation between the cosine of phrases and human-rated score\footnote{Human scores are available in the datasets}. In this case, we could determine the systematic compositionality of each model by measuring the ability in capturing compositional phrase information beyond lexical content.
    \item \underline{Adversarial Paraphrase Sentences Classification}. It determines if the semantics hold when partial words in the input sentence are swapped. To achieve this, we fine-tune each model on PAWS and test them on an adversarial PAWS dataset, called PAWS (swap-only) \cite{zhang2019paws}. PAWS (swap-only) simulates the changing of word compositions (orderings). It contains both paraphrase and non-paraphrase pairs with high bag-of-words overlap and word swapping. In this study, we report the accuracy of the successful identification of non-paraphrase pairs to assess our effectiveness.
\end{itemize}

\textbf{Main Results}. As shown in Table \ref{tab:system composition}, the results suggest that \textit{CAT$_{mix}$ and CAT$_{comp}$ show superiority on systematic compositionality on both tasks}. On average, CAT$_{mix}$ improves the performance on the adversarial paraphrase sentences classification by 1.10 and phrase correlation by 6.42 compared to our primary baseline BERT. Additionally, it enhances the performance on the adversarial paraphrase sentences classification by 0.26 and phrase correlation by 5.22 compared to the best baseline (i.e., RoBERTa). These observations imply that after being pre-trained to compose primitives, CAT improves the ability in identifying the semantics correlation formed by word compositionalities. Although CAT$_{comp}$ is notably superior to CAT$_{mix}$, we would demonstrate in Sections \ref{q3} that CAT$_{mix}$ strikes a good balance between systematic compositionality and effectiveness on standardized language understanding task.

In detail, regarding the adversarial paraphrase sentences classification, CAT$_{comp}$ achieves the best results thanks to the explicit systematic compositionality modeling, which improves its performance by 2.11 compared to BERT, 1.27 compared to RoBERTa, and 2.89 compared to DistilBERT. Moreover, CAT$_{cont}$ performs similarly to BERT, while CAT$_{mix}$ outperforms BERT by 1.10. Interestingly, despite RoBERTa using more pre-training data than BERT and ours, it is still weaker than CAT$_{comp}$ and CAT$_{mix}$. These findings highlight the significance of systematic compositionality modeling in enhancing the performance of pre-trained models on compositionality-aware phrasal tasks.

Regarding the phrase correlation, CAT$_{mix}$ performs significantly better than BERT and even outperforms the best baseline, RoBERTa, by 7.4 on BiRD and 3.04 on BiRD-ABBA. However, we also observe that the overall results on BiRD-ABBA are much lower than BiRD, as identifying the relatedness of bigram pairs with high word overlap is more challenging. In fact, it requires assessing semantic changes in different compositions of the same group of words. We further tested ChatGPT (cf. Appendix \ref{asd8fonq34wrf}), which has a superabundance of training data and several hundred times more parameters than BERT. We found that its score on BiRD-ABBA is very unsatisfying. In particular, the result of ChatGPT scoring on BiRD data is 45.65, while the result on BiRD-ABBA plummets to 23.28. This indicates poor alignment with human judgments of phrase meaning similarity. These results suggest that, at least on this compositionality-aware task, ChatGPT, which may seem "omnipotent", may be far from the general intelligence demonstrated in humans.

\noindent\textbf{5.2.2 Evaluation on Out-of-distribution Datasets}

\textbf{Task \& Dataset}. After pre-training on corpora with diverse words and their combinations, we aim to evaluate the compositional generalization of each model to out-of-distribution (OOD) datasets. To achieve this, our evaluation follows \citet{hendrycks2020pretrained}. Given a dataset pair $(A, B)$, we fine-tune each model on $A$ (i.e., IID), and test it on $B$ (i.e., OOD) that contains realistic distribution shifts to $A$. Following \citet{hendrycks2020pretrained}, we consider the following dataset pairs: IMBD \cite{maas2011learning} and SST2 \cite{socher2013recursive}, STSB-image and STSB-MSRvid \cite{cer2017semeval}, MNLI-telephone and MNLI-letters \cite{williams2018broad}, AMAZON-music and AMAZON-video \cite{he2016ups}. For STSB, we report Pearson’s correlation coefficient, while for the other datasets, we report accuracy. See Appendix \ref{system_apa} for more information on the datasets and model fine-tuning.

\textbf{Main Results}. As shown in Table \ref{tab:system ood}, \textit{CAT$_{mix}$ is better at generalizing to out-of-distribution composed semantics}. Due to pre-training on a larger corpus, RoBERTa demonstrates significant performance in general language understanding tasks (i.e., IID) compared to others. However, despite the IID effectiveness of RoBERTa, it suffers a significant decrease in effectiveness on OOD datasets. On the contrary, CAT$_{mix}$ exhibits advantages in OOD data, with an average improvement of 1.85 compared to RoBERTA (3.89 on SST2, 1.09 on STSB-MSRvid, 1.02 on MNLI-Latter, and 1.38 on AMAZON-Video) and 1.83 compared to BERT (2.75 on SST2, 1.13 on STSB-MSRvid, 1.17 on MNLI-Latter, and 2.28 on AMAZON-Video). 

Compared to CAT with different representations, CAT$_{cont}$ is better on IID datasets than CAT$_{comp}$, with an average improvement of 1.45. This empirical evidence is consistent with the previous conclusion of theoretical analysis \cite{liu2021discrete}, stating that the combinatorial expressiveness of discrete codes is able to model complex language phenomena, but it is still weaker than the contextual representations. However, after mixing CAT$_{cont}$ and CAT$_{comp}$ together, CAT$_{mix}$ strikes a good balance.

\subsection{Robustness of Systematic Compositionality}
\label{q2}
CAT benefits from the discreteness of primitives and their composition. Research conducted earlier has demonstrated that discrete variables possess the attribute of being robust to noise \cite{liu2021discrete}. This finding has inspired us to conduct a thorough examination and evaluate the robustness of CAT in the process of composing discrete primitives. Specifically, we are interested in examining the multi-primitive composition module in CAT, which filters out irrelevant codes/primitives from the compositions of $h_{comp}$ and enables us to break down $h_{cont}$ into several dynamic-sized latent codes. In this section, our objective is to determine whether CAT is capable of filtering out irrelevant primitives during composition.

\textbf{Task \& Dataset}. Our goal is to determine if CAT can filter out irrelevant primitives during composition. For this purpose, following \citet{jia-liang-2017-adversarial}, we test the ability of CAT to comprehend contexts that include adversarially inserted irrelevant sentences and answer questions about the given contexts. We utilize reading comprehension datasets, i.e., SQuAD \cite{rajpurkar2016squad} and its variant with adversarial noises. These noises are automatically generated to distract models without altering the correct answer or misleading humans. Essentially, we fine-tune each model on the SQuAD and test it on the SQuAD-adversarial. See Appendix \ref{discre_apa} for more details.

\begin{figure}[!htb]
    \centering
    \includegraphics[width=0.45\textwidth]{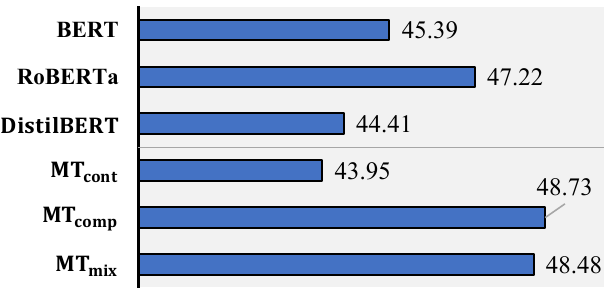}
    \caption{Illustration of robustness evaluation. Here, we report the accuracy of each model.}
    \label{fig:robust}
\end{figure}

\textbf{Main Results.} As illustrated in Figure \ref{fig:robust}, our CAT, which employs mixed representations, exhibits remarkable robustness to adversarial samples. It outperforms BERT by 3.09 and the best baseline by 1.26. This performance gain may be attributed to the fact that semantic decomposition based on a discrete codebook is beneficial for filtering out irrelevant information in the [CLS] embedding. Such semantic decomposition improves the effectiveness of the compositional representation CAT$_{comp}$, which achieves even better performance than CAT$_{mix}$. One possible explanation is that irrelevant information contained in the contextual representation CAT$_{cont}$ may be fused into the CAT$_{mix}$ during our mixing procedure. However, this does not affect our advantage.


\subsection{Effectiveness on Standardized Test}
\label{q3}

Our previous study suggests the combinatorial effectiveness of discrete primitives \cite{liu2021discrete}. After demonstrating CAT's efficacy in compositionality-aware tasks, in this section, we proceed to assess the effectiveness of composed primitives in standardized tests. Our objective is to investigate whether the contextual information required by downstream tasks would be affected when the vanilla Transformer is equipped with systematic compositionality capabilities.

\textbf{Task \& Dataset}. In this section, we assess the language understanding capability on GLUE \cite{wang2018glue} following BERT, and report scores on each task after fine-tuning. In line with BERT, we report the F1 score for MRPC and QQP datasets, the Spearman correlation score for STSB, and the accuracy score for the remaining tasks. See Appendix \ref{discre_apa} for more details.

\textbf{Main Results}. According to Table \ref{tab:glue}, on average, CAT$_{cont}$ and CAT$_{mix}$ exhibit slightly better performance than BERT (+0.53, +0.09, respectively), while CAT$_{comp}$ is weaker. As previously discussed, the compositional representation CAT$_{comp}$ is better tailored for compositionality-aware and discretization-oriented tasks, while the contextual representation CAT$_{cont}$ is more appropriate for standardized tasks. More important, CAT$_{mix}$ strikes a good balance between compositionality-aware and standardized tasks. Our findings highlight that the proposed CAT is adept at capturing semantic compositionality and significantly outperforms baselines on compositionality-aware tasks, with a minimal impact on the effectiveness of standardized language understanding tasks.

\begin{table}[]
\centering
\caption{Effectiveness on standardized test. CAT has minimal impact on the effectiveness of the GLUE task}
\label{tab:glue}
\resizebox{0.45\textwidth}{!}{%
\begin{tabular}{l|c|c|c|c|c|c}
\toprule
\textbf{} & \multicolumn{1}{c|}{\textbf{BERT}} & \multicolumn{1}{l|}{\textbf{RoBERTa}} & \multicolumn{1}{l|}{\textbf{DistilBERT}} & \multicolumn{1}{l|}{\textbf{CAT$_{cont}$}} & \multicolumn{1}{l|}{\textbf{CAT$_{disc}$}} & \textbf{CAT$_{mix}$} \\ \midrule
CoLA      & 52.10                              & 61.60                                 & 47.60                                    & 60.18                                  & 50.09                                  & 60.28            \\
MNLI\_m   & 78.12                              & 80.12                                 & 73.62                                    & 75.28                                  & 72.85                                  & 74.28            \\
MNLI\_mm & 78.41                              & 79.41                                 & 74.61                                    & 75.97                                  & 74.40                                  & 75.19            \\
MRPC      & 74.40                              & 75.70                                 & 73.20                                    & 74.95                                  & 74.03                                  & 74.93            \\
QNLI      & 85.45                              & 87.75                                 & 82.85                                    & 84.21                                  & 81.19                                  & 83.21            \\
QQP       & 71.20                              & 77.90                                 & 70.10                                    & 76.19                                  & 71.78                                  & 76.23            \\
RTE       & 64.62                              & 65.92                                 & 55.22                                    & 62.93                                  & 60.68                                  & 62.76            \\
SST2      & 94.50                              & 95.80                                 & 93.10                                    & 93.98                                  & 93.44                                  & 94.14            \\
STSB      & 85.80                              & 87.29                                 & 83.70                                    & 85.71                                  & 82.85                                  & 84.42            \\ \midrule
\textbf{Avg.}      & 76.07                              & \textbf{79.05}                                 & 72.67                                    & \underline{76.60}                                  & 73.48                                  & 76.16           \\ \bottomrule
\end{tabular}%
}
\end{table}

\section{Conclusion}
Our research delves into the characteristics of systematic compositionality in human languages. For the first time, we propose a compositionality-aware Transformer (CAT) and two new pre-training tasks to facilitate systematic compositionality. We tentatively provide a successful implementation for multi-layer CAT and empirically verify its effectiveness. Our approach captures semantic compositionality better and significantly outperforms baselines on compositionality-aware tasks, with minimal impact on the effectiveness of standardized language understanding tasks.

Systematic compositionality is widely believed to be characteristic of human intelligence. Our study provides a primary exploration of this challenge, and our findings may provide an important proof-of-concept for producing better LLMs, which crystallizes the past experiences and generalizes them to the new composed contexts.

\section*{Acknowledgement}
This work was supported in part by the National Natural Science Foundation of China (No. 62272330); in part by the Fundamental Research Funds
for the Central Universities (No. YJ202219).

\bibliography{aaai24}
\clearpage
\appendix

\section{Implementation Details}
\label{ideital}
\subsection{Implementation of multi-layer CAT}
\textbf{Model Architecture}. To work within the limitations of our computation resources, we utilized a 4-layer CAT model in our experiments (referred to as MCAT). This model has a hidden size of 768, 12 attention heads, and a sequence length of 128. Each block, except for the final one, contains a codebook with 1000 codes, where each code is a 768-dimensional vector. The last block's codebook is the same size as Sememes, containing 3152 codes. Our implementation of the model was done in Pytorch, with the primary code referencing BERT-Python\footnote{\url{ https://github.com/codertimo/BERT-pytorch}}. We used the open-source code\footnote{\url{https://github.com/rishikksh20/rectified-linear-attention}} from the paper on sparse attention \cite{zhang2021sparse} for the CAT.


\textbf{Input/Output Representations}. Our input representation is identical to BERT's, and it can be implemented effortlessly using the Huggingface library. However, unlike BERT's output, which only produces a contextual representation, our CAT expands it into three types: contextual, compositional, and mixed representations. This means, it brings more flexibility to the downstream applications.

\subsection{Implementation of baselines}
To ensure a fair comparison, we utilized the Huggingface implementations of RoBERTa and DistilBERT. The only difference between these two models in our experiments is that they are also 4-layer-based, with a hidden size of 768 and a sequence length of 128.

\section{Appendix for Model Pre-training}
\label{pre-trainingt}

\subsection{Pre-training Tasks}
\label{pre-trainingtask}
Four pre-training tasks are included in our paper, i.e., mask LM, next sentence prediction, guided decomposition, and semantics composition tasks. Here, we elaborate on the last two tasks and refer readers to \citet{kenton2019bert} for details on the first two tasks. 

\textbf{Guided Decomposition Task}. To enhance the optimization of the sparse attention mechanism in CAT (or the last layer of the multi-layer CAT), we utilized sememes in OpenHowNet to guide the mechanism to attend to the appropriate sememes in the codebook. Here, $x_i$ represents the $i$-th word in an input sentence of length $N$, $e \in R^{K \times m}$ is the codebook in CAT (or the last layer of the multi-layer CAT), $O_i \in R^K$ is the attention weight to the codebook and $s_i \in R^K$ is a $\{0,1\}$ vector containing the correct sememes for $x_i$, where $s_{ij}=1$ if the $j$-th sememe belongs to $x_i$, otherwise $s_{ij}=0$. To calculate the loss $L_{GD}$ for incorrect sememe-matchings, we averaged the attention weights that are focused on the wrong sememes. Furthermore, we introduced the L1 norm on the attention weights from all layers to encourage sparsity.

$$L_{GD}(x_i) = \sum_{x_i}\frac{sum((1-s_i)\odot O_i)}{sum(1-s_i)} + L_1(O_i)$$

There are only 3152 sememes in the OpenHowNet, which cover a mere 104027 Chinese words (and their corresponding English translations). This represents a small portion of our pre-training corpus, requiring the model to learn and generalize efficiently from only a few supervisions.

\textbf{Semantics Composition Task}.
We involve three learning objectives, including the \textit{Reconstruction Consistency}, \textit{Semantic Sufficiency}, and \textit{Nuance Minimization}. To ensure the similarity between contextual and compositional representations in the vector and semantic spaces, we utilized two learning objectives.
\begin{itemize}
    \item \textit{Reconstruction Consistency} aims to encourage similarity in the vector space. To achieve this, we used the stop gradient operator $sg$ and a hyper-parameter $\beta=0.1$ to focus more on updating the compositional representation. In this case, it shifts focus more on updating compositional representation, rather than the contextual one.  
    \begin{equation}
        \ell_{rc} = \| h_{comp} - sg(h_{cont}) \|_2^2 + \beta \| sg(h_{comp}) - h_{cont} \|_2^2
    \end{equation}

    \item \textit{Semantic Sufficiency} aims to encourage similarity in the semantic space. This learning objective assumes that both compositional and contextual representation should be effective in the Masked LM and Next Sentence Prediction tasks. 
    \begin{equation}
    \begin{split}
        \ell_{ss} &= MLM(h_{comp}) + MLM(h_{cont}) \\
                  &+ NSP(h_{comp}) + NSP(h_{cont})\\
                  &+ NSP(h_{mix}) + NSP(h_{mix})
    \end{split}
    \end{equation}
    
    \item \textit{Nuance Minimization} aims to ensure that the difference between contextual and compositional vector representations $H_{nu} = h_{cont} - h_{comp}$ is insignificant and less informative to the main semantics. To achieve this, we maximize the entropy of $H_{nu}$ and minimize its performance on the Masked LM and Next Sentence Prediction tasks.
    \begin{equation}
\begin{split}
       \ell_{nm} &= exp(-Entropy(H_{nu})) \\
                 &+ exp(-MLM(h_{cont})) + exp(-NSP(h_{comp}))
\end{split}
    \end{equation}
\end{itemize}


The loss function for the semantics composition task can be summarized as $L_{SC}=\ell_{rc} + \ell_{ss} + \ell_{nm}$. In the case of CAT pre-training, the overall loss function is defined as $L = L_{task} + L_{GD} + L_{SC}$. Here, $L_{task}$ refers to the loss incurred by the contextual representation for the next sentence prediction and masked LM tasks.

\subsection{Pre-training Procedure}
\label{pre-trainingd}
Our pre-training procedure is based on the existing literature on language model pre-training, with the addition of two novel pre-training tasks: the guided decomposition task and the semantics composition task. The former supervises the decomposition process, while the latter controls the compositional semantics of discrete codes containing sufficient semantics during discretization and composition. For pre-training, we utilized the BERT dataset and OpenHowNet for semantic grounding supervision. Additional details on our pre-training tasks can be found in Appendix \ref{pre-trainingtask}.

To carry out pre-training, we followed the BERT pre-training procedure and utilized NVIDIA RTX A6000 GPUs with Distributed Data-Parallel. Our pre-training data consisted of BooksCorpus (800M words) and English Wikipedia (2,500M words). For more information, please refer to the BERT paper (cf. Pre-training Procedure Section in \cite{kenton2019bert}). Note that we pre-trained our model with a sequence length of 128 for all steps to save time.

\begin{table*}[!thb]
\centering
\caption{PAWS-QQP and PAWS (swap-only) Examples}
\label{tab:PAWS-QQP}
\resizebox{\textwidth}{!}{%
\begin{tabular}{l|l|l}
\toprule
\multicolumn{1}{l|}{\textbf{Sentence 1}} & \multicolumn{1}{l|}{\textbf{Sentence 2}} & \textbf{Label} \\ \midrule
\multicolumn{3}{c}{\textbf{PAWS-QQP}} \\ \midrule
\multicolumn{1}{l|}{{ \begin{tabular}[c]{@{}l@{}}In Paris, in October 1560, \\ he secretly met the English ambassador, \\ Nicolas Throckmorton, \\ asking him for a passport to return to England through Scotland.\end{tabular}}} & \multicolumn{1}{l|}{{\begin{tabular}[c]{@{}l@{}}In October 1560,\\ he secretly met with the English ambassador, \\ Nicolas Throckmorton, in Paris, \\ and asked him for a passport to return to Scotland through England.\end{tabular}}} & { 0} \\ \midrule
\multicolumn{1}{l|}{\begin{tabular}[c]{@{}l@{}}The NBA season of 1975 -- \\ 76 was the 30th season of the National Basketball Association\end{tabular}} & \multicolumn{1}{l|}{\begin{tabular}[c]{@{}l@{}}The 1975 -- \\ 76 season of the National Basketball Association was the 30th season of the NBA\end{tabular}} & 1 \\ \midrule
\multicolumn{3}{c}{\textbf{PAWS (swap-only)}} \\ \midrule
\multicolumn{1}{l|}{The Liberal Party regained South Coast but lost the outer Sydney seat of Camden} & \multicolumn{1}{l|}{The Liberal Party lost South Coast but regained the outer Sydney seat of Camden} & 0 \\ \midrule
\multicolumn{1}{l|}{\begin{tabular}[c]{@{}l@{}}The AVA includes about in Iredell, Wilkes and Yadkin counties.\\ The designation, the second in North Carolina, took effect May 27, 2008.\end{tabular}} & \multicolumn{1}{l|}{\begin{tabular}[c]{@{}l@{}}The AVA includes about in Yadkin, Wilkes and Iredell counties. \\ The designation, the second in North Carolina, took effect May 27, 2008.\end{tabular}} & 1 \\ \toprule
\end{tabular}%
}
\end{table*}

\section{Appendix for Evaluation Tasks}
\label{eeeeeeee}
\subsection{Systematic Compositionality Evaluation}
\label{system_apa}
Based on recent works \cite{yu2021interplay, yu2020assessing}, we conducted an evaluation of phrasal representation in transformers by fine-tuning the model on datasets that require composition. This evaluation included two tasks: Phrase Correlation and Adversarial Paraphrase Sentences Classification. We also consider the compositional generation to out-of-distribution scenarios. 

\textbf{Phrase Correlation}. Following the approach of \citet{yu2021interplay, yu2020assessing}, we evaluate whether our method and baselines capture compositional phrase information beyond lexical content. We fine-tune each model on the adversarial paraphrase dataset called PAWS-QQP \cite{zhang2019paws} and have each model classify paraphrases with high lexical overlap. Next, we test each fine-tuned model on the BiRD and BiRD-ABBA datasets \cite{asaadi-etal-2019-big}.

\begin{itemize}
    \item \underline{Dataset.} To fine-tune our model, we used the corresponding dataset in Huggingface for \textit{PAWS-QQP}, which consists of sentence pairs with high lexical overlap and is formulated as a binary classification of whether two sentences are paraphrases or not. Examples can be found in Table \ref{tab:PAWS-QQP}. For evaluating our model's alignment with human judgments of phrase meaning similarity, we use the \textit{BiRD} dataset, which contains over three thousand bigram pairs with human-annotated similarity scores ranging from 0 to 1. We also utilized \textit{BiRD-ABBA}, which contains 410 "AB-BA" mirror-image pairs with 100\% word overlap (e.g., law school/school law). This dataset is more challenging as it requires assessing semantic changes in different compositions of the same group of words. Examples of both datasets can be found in Table \ref{tab:example222}. 
    \item \underline{Metrics}. To assess the effectiveness of our models at capturing compositional phrase information beyond lexical content, we follow \cite{yu2021interplay, yu2020assessing} and report the Pearson correlation between cosine of phrases and human-rated score for all source-target pairs in both BiRD and BiRD-ABBA. The cosine of an input pair is calculated based on the [CLS] embeddings of the two phrases.
\end{itemize}

\begin{table}[!htb]
\centering
\caption{Compositionality examples on BiRD and BiRD-ABBA \cite{asaadi-etal-2019-big}.}
\label{tab:example222}
\resizebox{0.4\textwidth}{!}{%
\begin{tabular}{cc|c}
\toprule
\multicolumn{1}{l|}{\textbf{Phrase 1}} & \multicolumn{1}{l|}{\textbf{Phrase 2}} & \textbf{Semantics} \\ \hline
\multicolumn{3}{c}{\textbf{BiRD}} \\ \hline
\multicolumn{1}{l|}{{Access information}} & \multicolumn{1}{l|}{{Auditor general}} & {0.293} \\ \hline
\multicolumn{1}{l|}{Adult male} & \multicolumn{1}{l|}{Men} & 0.844 \\ \hline
\multicolumn{3}{c}{\textbf{BiRD-ABBA}} \\ \hline

\multicolumn{1}{l|}{{Data processing}} & \multicolumn{1}{l|}{{Processing data}} & {0.827} \\ \hline
\multicolumn{1}{l|}{Data point} & \multicolumn{1}{l|}{Point data} & 0.398 \\\toprule
\end{tabular}%
}
\end{table}

\textbf{Adversarial Paraphrase Sentences Classification}. Our evaluation aims to determine if the semantics hold even when the input sentences are disturbed. To achieve this, we fine-tuned each model on the PAWS-QQP dataset and tested them on an adversarial PAWS dataset, called PAWS (swap-only) \cite{zhang2019paws}, which evaluates the compositional phrases with partial word swaps. 

\begin{itemize}
    \item \underline{Dataset.} For fine-tuning, we utilized the same PAWS-QQP dataset for each model. To evaluate our models, we used the PAWS (swap-only) dataset, which contains both paraphrase and non-paraphrase pairs with high bag-of-words overlap and word swapping. We utilized the dataset in Huggingface and used its training data for the Adversarial Paraphrase Sentences Classification testing. Examples can be found in Table \ref{tab:PAWS-QQP}. 
    \item \underline{Metrics}. Our effectiveness is assessed by reporting the accuracy of successfully identifying non-paraphrase pairs in this study. 
\end{itemize}

\textbf{Out-of-distribution generalization}. Our goal is to assess the ability of each model to generalize to out-of-distribution (OOD) datasets, following \citet{hendrycks2020pretrained}. To achieve this, we fine-tuned each model on an in-distribution (IID) dataset, $A$, and tested it on an OOD dataset, $B$, that contains realistic distribution shifts to $A$. 

\begin{itemize}
    \item \underline{Dataset}. We used the following dataset pairs, all from Huggingface, as suggested by \citet{hendrycks2020pretrained}: IMBD \cite{maas2011learning} and SST2 \cite{socher2013recursive}, STSB-image and STSB-MSRvid \cite{cer2017semeval}, MNLI-telephone and MNLI-letters \cite{williams2018broad}, AMAZON-music and AMAZON-video \cite{he2016ups}. 
    \begin{itemize}
        \item \textit{IMBD \& SST2}. We utilized the full-length lay movie reviews in IMBD and the pithy expert movie reviews in SST2. As IMBD on Huggingface does not have a labeled validation dataset, we split the IMBD test dataset into two folds (1:1) as the validation and test datasets, respectively. Additionally, we used the validation of SST2 for OOD testing. 
        \item \textit{STSB-image \& STSB-MSRvid}. The STSB dataset requires predicting the semantic similarity between pairs of sentences and contains the text of different genres and sources. Here, we resort to image and MSRvid. In our experiments, we use the established training, validation, and test datasets in Huggingface.
        \item \textit{MNLI-telephone \& MNLI-letters}. MNLI is a textual entailment dataset using sentence pairs drawn from different genres of text. We follow previous studies and select examples from two genres of transcribed text (i.e., telephone) and one genre of written text (i.e., letters). We randomly select 80,000 samples from the training dataset of MNLI-telephone in Huggingface, with the remaining samples serving as the validation dataset. The validation datasets of MNLI-telephone in Huggingface are used as the IID test dataset, while the validation dataset of MNLI-letters is used as the OOD test dataset.
        
        \item \textit{AMAZON-music \& AMAZON-video}. We use two categories of entertainment products (i.e., music and video) from the Amazon Review Dataset, which contains product reviews from Amazon, and each model is required to predict a review's 1 to 5-star rating. As Huggingface only contains labeled training datasets of the Amazon Review Dataset, we randomly select 50,000 instances from AMAZON-music as training data, 10,000 instances for validation, and 5,000 for testing. For AMAZON-video, 5,000 instances are also randomly selected as the OOD test dataset. 
    \end{itemize}
    \item \underline{Metrics}. We report the Pearson's correlation coefficient for STSB and accuracy for the other datasets.
\end{itemize}

\subsection{Characteristics of Composed Primitives}
\label{discre_apa}
Previous works have shown that discrete variables/representations exhibit the characteristics of \romannumeral1) being robust to noise \cite{liu2021discrete} \romannumeral2) and having sample efficiency \cite{chaabouni2020compositionality, goyal2019recurrent, bengio2017consciousness}, and 3) finally its effectiveness on standardized tasks. In this study, we aim to verify whether these characteristics hold in our compositional representations.

\textbf{Noise Robustness Evaluation}. To test the ability of systems to answer questions about paragraphs containing adversarially inserted noisy sentences, we follow \citet{jia-liang-2017-adversarial} and utilize the SQuAD dataset \cite{rajpurkar2016squad} with adversarial noise. We fine-tune each model on the vanilla SQuAD dataset and test them on the SQuAD-adversarial dataset. 
\begin{itemize}
    \item \underline{Dataset}. \textit{SQuAD} is a reading comprehension dataset, consisting of questions posed by crowd workers on a set of Wikipedia articles, where the answer to every question is a segment of text, or span, from the corresponding reading passage, or the question might be unanswerable. \textit{SQuAD-adversarial} contains adversarially inserted sentences that are automatically generated to distract computer systems without changing the correct answer or misleading humans. Both datasets are available in Huggingface, and we utilized the training and validation sets of SQuAD for model fine-tuning and the validation set of SQuAD-adversarial for testing.
    \item \underline{Metrics}. Our evaluation of each model is based on standard accuracy metrics.
\end{itemize}

\begin{figure}[!htb]
\centering
\begin{tabular}{c}
\includegraphics[width=0.35\textwidth]{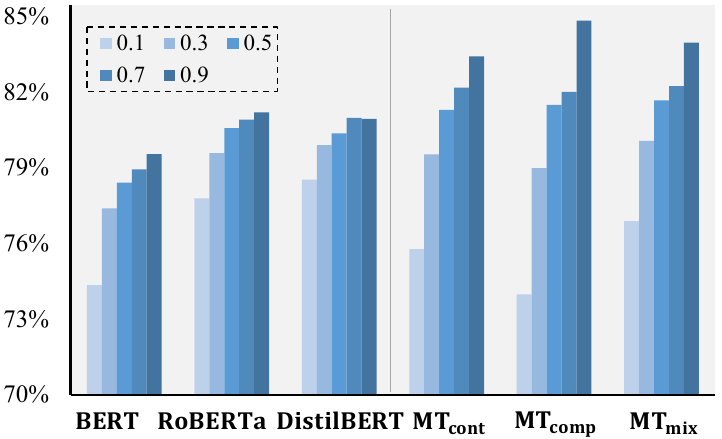}\\
(a) IMDB\\
\includegraphics[width=0.35\textwidth]{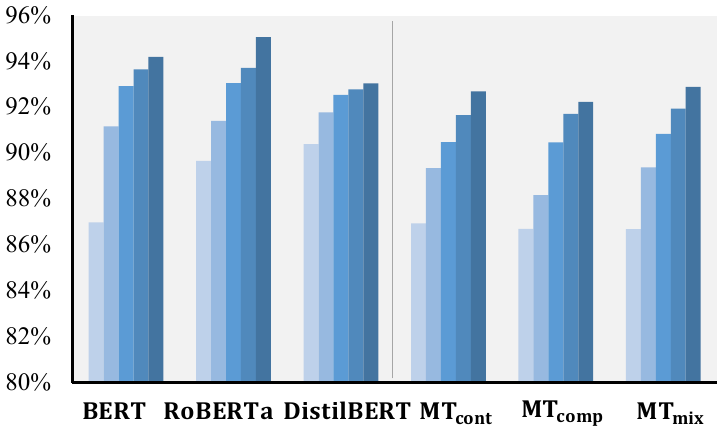}\\
(b) SST2
\end{tabular}
\caption {Illustration of sample efficiency evaluation.}

\label{fig:sample}

\end{figure}

\begin{table}[!htb]
\centering
\caption{Average performance gain under the same sample size increment }
\label{tab:size}
\resizebox{0.45\textwidth}{!}{%
\begin{tabular}{ccccccc}
\toprule
\textbf{Data} & \textbf{BERT} & \textbf{RoBERTa} & \textbf{DistilBERT} & \textbf{CAT$_{cont}$} & \textbf{CAT$_{comp}$} & \textbf{CAT$_{mix}$} \\ \midrule
IMDB & 1.30 & 0.85 & 0.60 & \underline{1.91} & \textbf{2.72} & 1.77 \\
SST2 & \textbf{1.80} & 1.35 & 0.66 & 1.44 & 1.38 & \underline{1.55} \\ \hline
\end{tabular}%
}
\end{table}

\textbf{Sample Efficiency Evaluation}. We compare the performance of various models on different amounts of training data to measure their sample efficiency. Inspired by \cite{chen2022pathologies}, we vary the ratio of training data in $\{10\%, 30\%, 50\%, 70\%, 90\%\}$ during the fine-tuning stage of each model and evaluate it on the same test dataset. 

\begin{itemize}
    \item \underline{Datasets}. For simplicity, we focused on two datasets: \textit{IMBD} \cite{maas2011learning} and \textit{SST2} \cite{socher2013recursive}. As IMDB on Huggingface does not contain a labeled validation set, we split the test dataset into two folds (1:1) to create validation and test sets, respectively. Additionally, SST2 on Huggingface does not have a labeled test set, so we randomly set aside 60,000 samples from the training set and regarded the remaining samples as the test set. 
    \item \underline{Metrics}. Here, we report the accuracy. 

    \item \underline{Main Results}. While previous studies have suggested that discrete codes/variables have higher sample efficiency \cite{liu2021discrete}, our experiment does not find any concrete evidence to support the notion that compositional representation derived from discrete codes has better sample efficiency (cf. Figure \ref{fig:sample} in the Appendix). In particular, Given limited samples, although we observe the CAT$_{mix}$'s improvement on IMDB compared to BERT, this did not hold true for SST2. Similarly, we found no evidence to suggest that CAT$_{comp}$ is superior to CAT$_{cont}$. To gain a clearer understanding, we calculate the average performance gain of each model under the same sample size increment (i.e., a 20\% increase), which is the average difference in performance between adjacent bars in Figure \ref{fig:sample}. We summarize the results in Table \ref{tab:size}, and discover that the average performance gain did not exhibit a constant advantage. One possible explanation is that combining discrete codes for downstream tasks still requires a considerable number of samples to learn.
\end{itemize}

\textbf{Effectiveness on Standardized Test}. We assess the language understanding capability on GLUE \cite{wang2018glue} and report scores on each task after fine-tuning. 

\begin{itemize}
    \item \underline{Datasets}. We refer the readers to \citet{wang2018glue} for details. We use the glue dataset on Huggingface in the experiments. 
    \item \underline{Metrics}. In line with BERT, we report the F1 score for MRPC and QQP datasets, Spearman correlation score for STSB, and accuracy score for the remaining tasks.
\end{itemize}

\subsection{Fine-tuning Details}
\label{fine-tuned}
For each task, we simply plug the task-specific inputs and outputs into each model and fine-tune the parameters end-to-end. During fine-tuning, we varied the learning rate (Adam) within $\{1e^{-5}, 2e^{-5}, 3e^{-5}, 4e^{-5}, 5e^{-5}\}$, batch size within $\{256, 128, 64, 32\}$, and epoch within $4$. We selected the best model based on its performance on the validation set for downstream task testing. All training data are randomly shuffled. It is worth noting that our CAT supports three fine-tuning modes: compositional representation, contextual representation, or both. Our experiments showed that the compositional representation is more suitable for compositionality-aware and discretization-oriented tasks, while the contextual representation is more tailored for general tasks like standardized language understanding. The integration of both representations achieves a trade-off between the two. 

\section{Appendix for ChatGPT Scoring}
\label{asd8fonq34wrf}
Inspired by \cite{yu2021interplay, yu2020assessing}, we assess the alignment with human judgments of phrase meaning similarity on BiRD \cite{asaadi-etal-2019-big}, a bigram relatedness dataset designed to evaluate the compositionalities. We also filter the BiRD dataset to pairs in which the two phrases consist of the same words, which we refer to as the BiRD-ABBA dataset. See dataset samples in Table \ref{tab:example222}.

As ChatGPT requires a fee and has been banned in some countries, we use Monica\footnote{\url{https://monica.im/}} for BiRD and BiRD-ABBA scoring, which is a Chrome extension powered by ChatGPT API. As free users of Monica, we have a daily usage limit. Thus, we require Monica to score multiple bigram pairs at a time (precisely, 20 at a time). We use the following prompt for scoring: \textit{Read the following word pairs and assign each one a score depending on how similar in meaning the word pair is. Each score should range from 0 to 1.} 

We found that the scores of ChatGPT on BiRD-ABBA are very unsatisfying. In particular, the result of ChatGPT scoring on BiRD data is 45.65, while the result on BiRD-ABBA plummets to 23.28. This indicates poor alignment with human judgments of phrase meaning similarity. These results suggest that, at least on this compositionality-aware phrasal task, ChatGPT, which may seem "omnipotent", may be far from the general intelligence demonstrated in humans.

\end{document}